\pdfoutput=1
\documentclass[11pt]{article}

\usepackage{ACL2023}
\usepackage{float}% Standard package includes
\usepackage{times}
\usepackage{latexsym}
\usepackage{booktabs}
\usepackage{graphicx}
\usepackage{amsmath}

\usepackage[T1]{fontenc}
\usepackage[utf8]{inputenc}

\usepackage{microtype}

\usepackage{inconsolata}
\title{Mao-Zedong At SemEval-2023 Task 4: Label Represention Multi-Head Attention Model With Contrastive Learning-Enhanced Nearest Neighbor Mechanism For Multi-Label Text Classification}

\author{Che Zhang\textsuperscript{+*} \and Ping'an Liu\textsuperscript{+} \and Zhenyang Xiao\textsuperscript{*} \and Haojun Fei\textsuperscript{+}\\
  \textsuperscript{[+]}Qifu Technology,China \\
  \textsuperscript{[*]}Peking University\\
  \texttt{{mmt,kjn}@stu.pku.edu.cn} \\
  \texttt{{liupingan-jk,zhangchulan-jk}@360shuke.com}
  }

\begin{document}
\maketitle
\begin{abstract}
The study of human values is essential in both practical and theoretical domains. With the development of computational linguistics, the creation of large-scale datasets has made it possible to automatically recognize human values accurately. SemEval 2023 Task 4\cite{kiesel:2023} provides a set of arguments and 20 types of human values that are implicitly expressed in each argument. In this paper, we present our team's solution. We use the Roberta\cite{liu_roberta_2019} model to obtain the word vector encoding of the document and propose a multi-head attention mechanism to establish connections between specific labels and semantic components. Furthermore, we use a contrastive learning-enhanced K-nearest neighbor mechanism\cite{su_contrastive_2022} to leverage existing instance information for prediction. Our approach achieved an F1 score of 0.533 on the test set and ranked fourth on the leaderboard. we make our code publicly available at \href{https://github.com/peterlau0626/semeval2023-task4-HumanValue}{https://github.com/peterlau0626/semeval2023-task4-HumanValue}.
\end{abstract}

\section{Introduction}

The identification and analysis of human values in texts has been an important area of research. With the development of computational linguistics, this research has gained widespread attention because of its potential impact on areas such as sentiment analysis, social science.

One of the challenges in this area is to accurately categorize all the human value. Several notable research achievements have been made in the categorization of human values. One of the approach is that classifies human values into 54 categories across four different levels\cite{kiesel:2022}. SemEval2023 task4 uses the classification method in this paper, where an argument is given to identify whether a value is included in the instrument, and the F1 scores of the results at the level2 level are used for the total ranking. There are 20 categories of human values in level 2, and a argument could belong to multiple value categories or not to any one value category. this is a typical multi-label text classification (MLTC) problem which has been applied in many scenarios such as news emotion analysis\cite{bhowmick_multi-label_2009} and web page tagging\cite{jain_extreme_2016}.

In this paper, we propose a model that combines the label-specific attention network with the contrastive learning-enhanced nearest neighbor mechanism\cite{su_contrastive_2022}. The multi-headed attention mechanism allows our model to overcome the shortcomings of traditional attention mechanism models and to be able to focus on different parts of a document, resulting in more accurate labeled attention results. And the nearest neighbor mechanism enables our model to not waste the rich knowledge that can be directly obtained from the existing training instances and helps enhance the interpretability and robustness of the model. 
\section{Background}
\subsection{Datasets}

The dataset comprises of arguments from six different domains such as news releases, online platforms, etc. originating from four different countries/regions, which are composed of 80\% data from IBM argument quality dataset (95\% from the original dataset), 15\% from the European Future Conference (New), and 5\% from group discussion ideas (2\% from the original dataset). The training dataset comprises of more than 6500 arguments, whereas the validation and test datasets consist of around 1500 arguments each. In addition, the organizers of the competition provided three additional datasets to evaluate the robustness of methods:validation set Zhihu (labels available), test set Nahj al-Balagha (labels confidential), test set The New York Times (labels confidential). All datasets have been manually annotated. 

Each sample in the dataset contains an argument ID, conclusion, stance towards the premise, and the premise itself. The labels consist of the argument ID and a column for each of the 20 value categories, indicating whether the sample belongs to each category (0 or 1).
\begin{figure}[htbp]
    \centering
    \includegraphics[width=6.5cm, height=4cm]{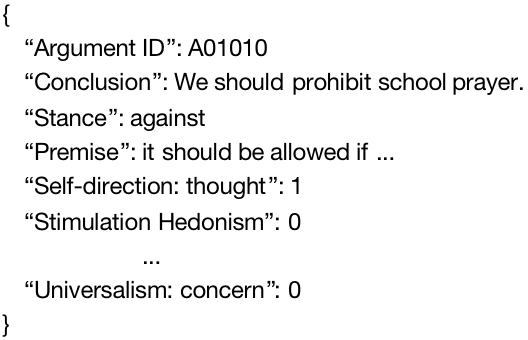}
\end{figure}

\subsection{Related Work}
 Before the widespread adoption of deep learning, models such as SVM were widely used to minimize an upper bound of the generalization error\cite{qin_study_2009}. Simple neural network (NN) models were later used for MLTC and achieved good performance\cite{nam_large-scale_2014}. Additionally, convolutional neural networks (CNNs) and recurrent networks with gated recurrent units (GRUs) have been successfully used with pre-trained word2vec embeddings\cite{berger_large_nodate}. Feature selection has been shown to be effective in speeding up learning and improving performance by identifying representative words and removing unimportant ones\cite{spolaor_evaluating_nodate}. 

In recent years, with the development of pre-trained models, the ability to extract semantic information has become increasingly powerful. there have been several representative works that have focused on improving the MLTC models. For example, \cite{pal_multi-label_2020} utilized graph neural networks based on label graphs to explicitly extract label-specific semantic components from documents. seq2seq model can capture the correlation between tags\cite{yang_sgm_2018}. LSAN\cite{xiao_label-specific_2019} can focus on different tokens when predicting each label.

\section{System Overview}
 
In this section, we will present our model, which consists of two main parts. The first part is a multi-headed attention mechanism based on a specific label representation, while the second part is a nearest neighbor mechanism enhanced using contrast learning.

The MLTC problem can be described as follows: assuming a set of data $D=\left\{\left(\mathrm{x}_{\mathrm{i}}, \mathrm{y}_{\mathrm{i}}\right)\right\}_{\mathrm{i}=1}^{\mathrm{N}}$, N labeled documents, where $x_i$ represents the text and $y_i \in\{0,1\}^l$ represents the label of $x_i$, and $l$ represents the total number of labels. Each document $x_i$ consists of a series of words. Our goal is to learn a classifier to establish a mapping from $x_i$ to $y_i$, so that when a new document $x$ is presented, its label vector y can be correctly predicted.As pretrained language models (PLMs) show remarkable performance in extracting natural language representations, we use PLMs as base encoder to get document and label feature. A input sample can be expressed as $\mathrm{x}_i=\left\{w_1, w_2, w_3 \ldots w_{n-1}, w_n\right\}$, $\mathrm{w}_{\mathrm{p}} \in R^{\mathrm{d}}$ denotes the $p$th word vector of a document. After calculated by PLMs , the input matrix of the whole sentence is obtained as $H \in R^{\mathrm{n} \times d}$, where d is the hidden dimension of PLMs.
\subsection {Label-specific multi-head attention network}

\begin{figure*}[h]
    \centering 
    \includegraphics[width=15.5cm,height=5cm]{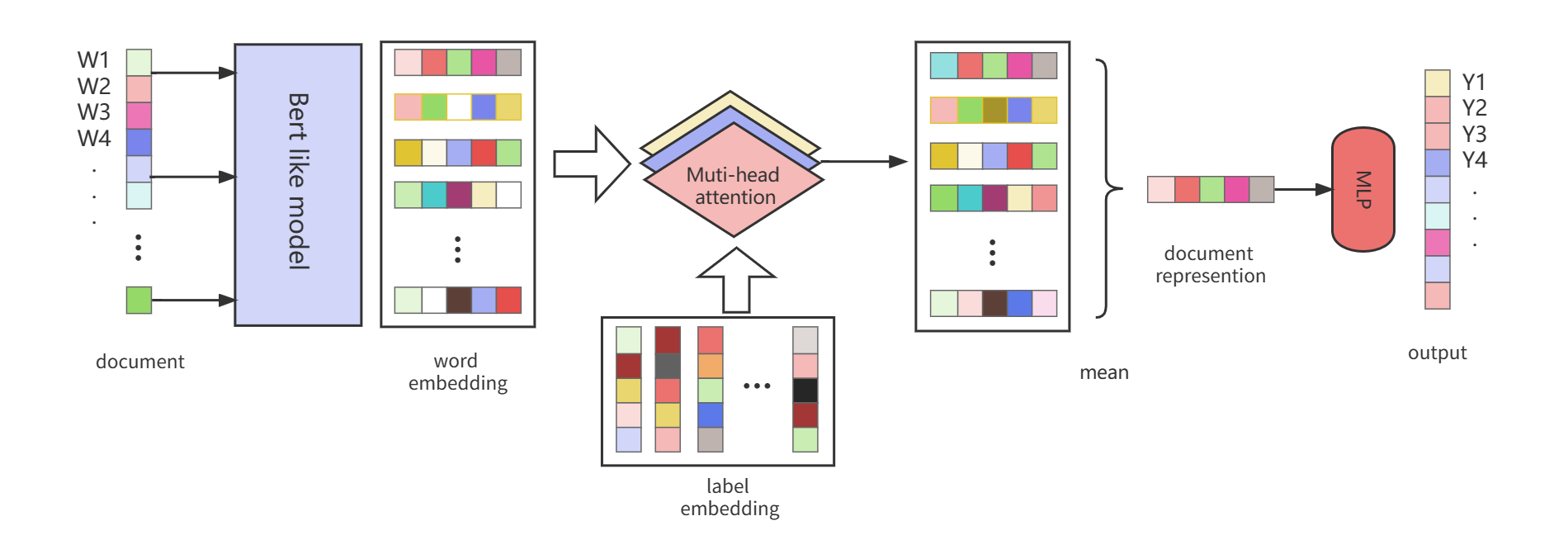}
    \caption{ The architecture of the proposed label-specific multi-attention network model}
    \label{1}
\end{figure*}

In order to explicitly capture the corresponding label-related semantic components from each document, the approach of using label-guided attention mechanisms to learn label-specific text representations has been widely used in previous studies, and such a method is used in LSAN\cite{xiao_label-specific_2019}. In addition, the success of the Transformer model\cite{vaswani_attention_2017} illustrates the ability of multi-headed attention mechanisms to extend the model's ability to focus on different locations more effectively than single-headed attention mechanisms. The usefulness of this method for text classification is very intuitive.For example in the following sentence, "Social media is good for us. Although it may make some people rude, social media makes our lives easier." Focusing on the words “although”, “but” and “makes life easier” at the same time is a more accurate way of getting at the value of comfort in life, while ignoring the the disadvantages of social media.As mentioned above, we next show our model.

Firstly, to make use of the semantic information of labels, we initialize the trainable label representation matrix $C \in R^{\mathrm{l} \times d}$ with the mean-pooling of the label features vector which is obtained by the pretrained encoder. Then, the multi-headed attention mechanism is used to compute the label-aware attention score. With the input document representation matrix as $H \in R^{\mathrm{n} \times d}$ and the label representation matrix $C$, the query $Q$, key $K$, and value $V$ of the attention mechanism can be expressed as follows:
\begin{equation}
\begin{aligned}
Q=W_{\mathrm{q}} C \\
K=W_{\mathrm{k}} H \\
V=W_v H
\end{aligned}
\end{equation}
Where $W_Q, W_K, W_V \in R^{d \times d}$ is the weight matrix to be learned.We use the h-head attention mechanism, then the three matrices $Q$,$K$,$V$ can be expressed in the following form.
\begin{equation}
\begin{aligned}
& \left(Q_1, Q_2, \ldots Q_h\right)=Q \\
& \left(K_1, K_2, \ldots K_h\right)=K \\
& \left(V_1, V_2, \ldots V_h\right)=V
\end{aligned}
\end{equation}
Where $Q_{\mathrm{i}} \in R^{l \times d_\mathrm{a}}, \quad K_{\mathrm{i}}, V_{\mathrm{i}} \in R^{\mathrm{n} \times d_\mathrm{a}}$ correspond to the query, key and value of each attention header, and ${ d_\mathrm{a}=d/h }$ denotes the dimensionality of a single attention mechanism representation space. Attention scores are then computed for each attention head similar to the method used in the Transformer model. Since the length of the document is different in a data batch, we perform a mask operation on the result of the QK matrix multiplication, set the value corresponding to the padding part to $1 e^{-12}$, and then use the softmax activation function to activate it.
\begin{equation}
\begin{aligned}
\operatorname{score}_{i}=\operatorname{softmax} \left(\operatorname{mask}\left(\frac{Q_{i} K_{i}}{d_{a}}\right)\right)
\end{aligned}
\end{equation}

$\operatorname{score}_{i} \in R^{1 \times n}$ denotes the attention score of the label for each word vector in the document. Then we obtain the attention results for each label with respect to the document content.
\begin{equation}
\begin{aligned}
Attention& =\\
\operatorname{Concat}&\left(\operatorname{attention}_1, \ldots, \operatorname{attention}_{\mathrm{h}}\right) W^O \\
\\
\text { where }& \operatorname{attention}_{\mathrm{i}} =\operatorname{score_i}\operatorname{V_i}
\end{aligned}
\end{equation}
Where $Attention \in R^{l \times d}$ can be considered as the representation vector of the document under the view of $L$ labels. To obtain the representation vector of the document $Z$, the row vectors of the Attention matrix for each labeled view are averaged:
\begin{equation}
\begin{aligned}
Z=\operatorname{mean}(\operatorname{row}(\operatorname{Attention}))
\end{aligned}
\end{equation}
After obtaining a comprehensive document representation with label-specific correlation, we can construct a multi-label text classifier by means of a perceptron consisting of fully connected layers. Mathematically, the predicted probability of each label of the next document can be determined by:
$\hat{\mathrm{y}}=\operatorname{sigmod} \left(W^1 Z^T\right)$.
Where $W^1 \in R^{l \times d}$is trainable parameters for the fully connected and output layers, which can transfer the output value into a probability. Since multi-label classification has the problem of unbalanced positive and negative samples, in order to balance the coefficients of positive and negative samples in the loss function and obtain a better trained model, we use the cross-entropy loss function with weights as the loss function of the model:
\begin{equation}
\begin{aligned}
L_{BCE}=\sum_{i=1}^b \sum_{\mathrm{j}=1}^l-(&\mathrm{w} \cdot \mathrm{y}_{\mathrm{ij}} \log \left(p_{i j}\right) \\ &+\left(1-y_{i j}\right) \log \left(1-p_{i j}\right))
\end{aligned}
\end{equation}
Where $b$ is the size of a data batch, $w$ is the weighting factor, $\mathrm{y}_{\mathrm{ij}}$ is the true value of the $j$th label of the $i$th sample, $\mathrm{p}_{\mathrm{ij}}$ is the probability that the model predicts that label to be $\mathrm{y}_{\mathrm{ij}}$, and $l$ is the total number of labels. The positive sample size/negative sample size in the training set is taken as the value of $w$.

\subsection {Contrastive Learning-Enhanced Nearest Neighbor Mechanism}

We use the k nearest neighbor (KNN) mechanism enhanced by contrast learning\cite{su_contrastive_2022}. This approach innovatively proposes a KNN mechanism for multi-label text classification that can make good use of the information of existing instances. And a contrastive learning approach is designed to enhance this KNN mechanism mechanism effectively. Specifically, this approach designs a loss function for contrastive learning based on dynamic coefficients of label similarity, which compares the documents representation vectors at training time to let the vector with more same labels be more similar as possible, while the vectors of documents with fewer identical labels are as far away as possible. Assuming a data batch of size $b$, we define a function to output all other instances of a particular instance in this batch $\mathrm{g}(\mathrm{i})=\{\mathrm{k} \mid \mathrm{k} \in\{1,2, \ldots, \mathrm{b}\}, \mathrm{k}\neq\mathrm{i}\}$.The contrastive loss(CL loss) of each instance pair (i, j) can be calculated as:
\begin{equation}
\begin{aligned}
&\mathrm{L}_{\mathrm{con}}^{\mathrm{ij}}=-\beta_{i j} \log \frac{e^{-d\left(z_i, z_j\right) / \tau^{\prime}}}{\sum_{k \in g(i)} e^{-d\left(z_i, z_k\right) / \tau^{\prime}}}\\
&C_{\mathrm{ij}}=y_i^T \cdot y_j, \quad \beta_{i j}=\frac{C_{i j}}{\sum_{k \in g(i)} C_{i k}}
\end{aligned}
\end{equation}
where $d(\cdot, \cdot)$ is the euclidean distance, $\tau^{\prime}$ is the contrastive learning temperature and $Z$ denotes the document representation. $C_{\mathrm{ij}}$denotes the label similarity between i, j, and normalized to obtain $\beta_{\mathrm{ij}}$. The CL loss of the whole batch can be expressed as:$L_{\text {con }}=\sum_i \sum_{j \in g(i)} L_{c o n}^{i j}$. The cross-entropy loss function is expressed as $L_{BCE}$, then the whole Loss function is $\mathrm{L}=L_{B C E}+\gamma L_{\text {con }}$, where a controls the ratio of the coefficients of the contrast learning loss function and the cross-entropy loss function. Then, we construct a data store of training instances so that we can later use the existing instance information as a comparison. Based on the training set $(x i, y i) \in D$, the storage of a training set of document representation vectors$D^{\prime}=\{(\text { hi, yi })\}_{i=1}^N$ is obtained by a trained model. where $h_{\mathrm{i}}$denotes the document representation vector of the training set, which is calculated by the model.
\begin{figure}[h]
    \centering 
    \includegraphics[width=7cm,height=5cm]{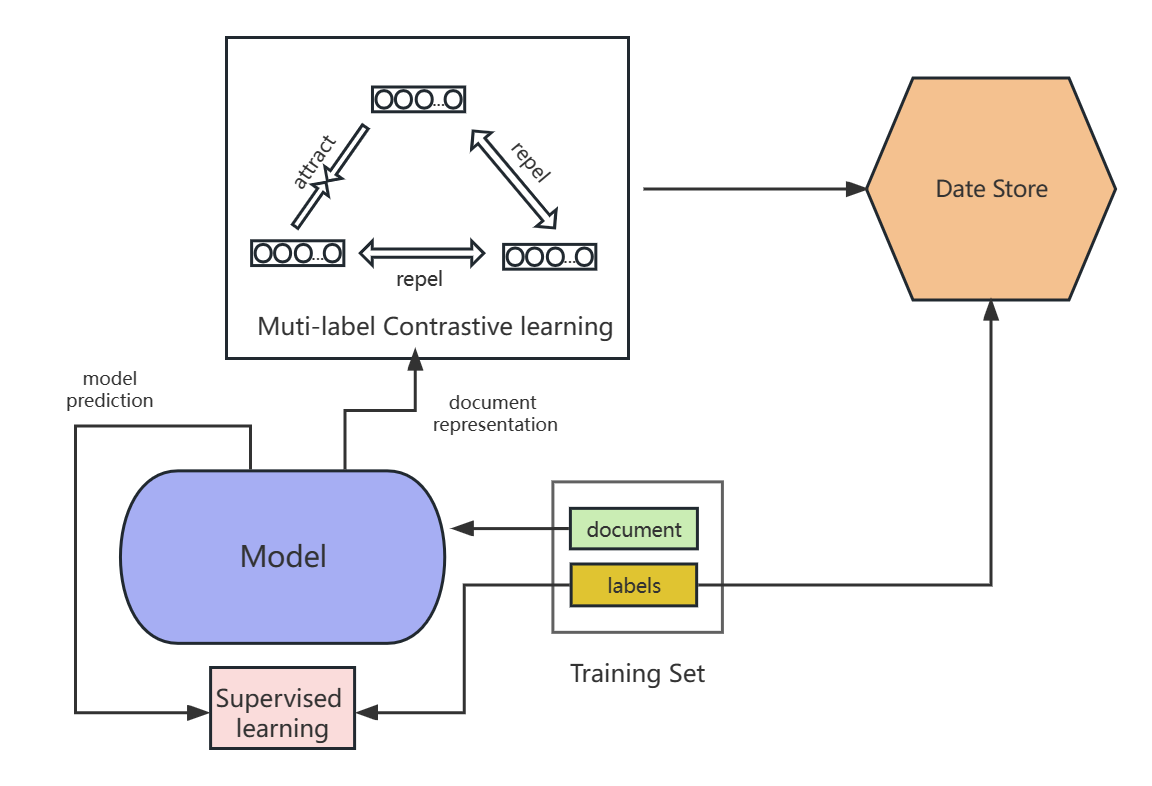}
    \caption{The training process of the wholel model with contrastive learning}
    \label{2}
\end{figure}

  In the inference stage, give an input $X$, after the model calculation we obtain its document representation vector $Z$, and the prediction of the model $\hat{y}_M \in\{\mathrm{p} \mid \mathrm{p} \in[0,1]\}^1$. Next,  we compare  with the data repository to find the nearest k nearest neighbors $\mathrm{N}=\{(\mathrm{hi}, \mathrm{yi})\}_{i=1}^{\mathrm{k}}$, then the KNN prediction can be calculated as:
\begin{equation}
\begin{aligned}
 & \hat{\mathrm{y}}_{K N N}=\sum_{i=1}^k \alpha_i y_i \\ & \alpha_i=\frac{e^{-d\left(h_i, Z\right)} / \tau}{\sum_j e^{-d(h j, Z)} / \tau}
\end{aligned}
\end{equation}
where $d(\cdot, \cdot)$ is the euclidean distance, $\tau$ is the temperature of KNN, $\alpha_i$ is the weight coefficient of the ith neighbor, when the closer the test instance vector representation is to this neighbor, the larger the weight will be. The final prediction form is expressed as follows:
\begin{equation}
\begin{aligned}
\hat{\mathrm{y}}=\lambda \hat{\mathrm{y}}_{K N N}+(1-\lambda) \hat{\mathrm{y}}_M
\end{aligned}
\end{equation}
where $\lambda$ is the weight coefficient that regulates the KNN prediction and the model prediction.
\begin{figure}[h]
    \centering 
    \includegraphics[width=6cm,height=5.5cm]{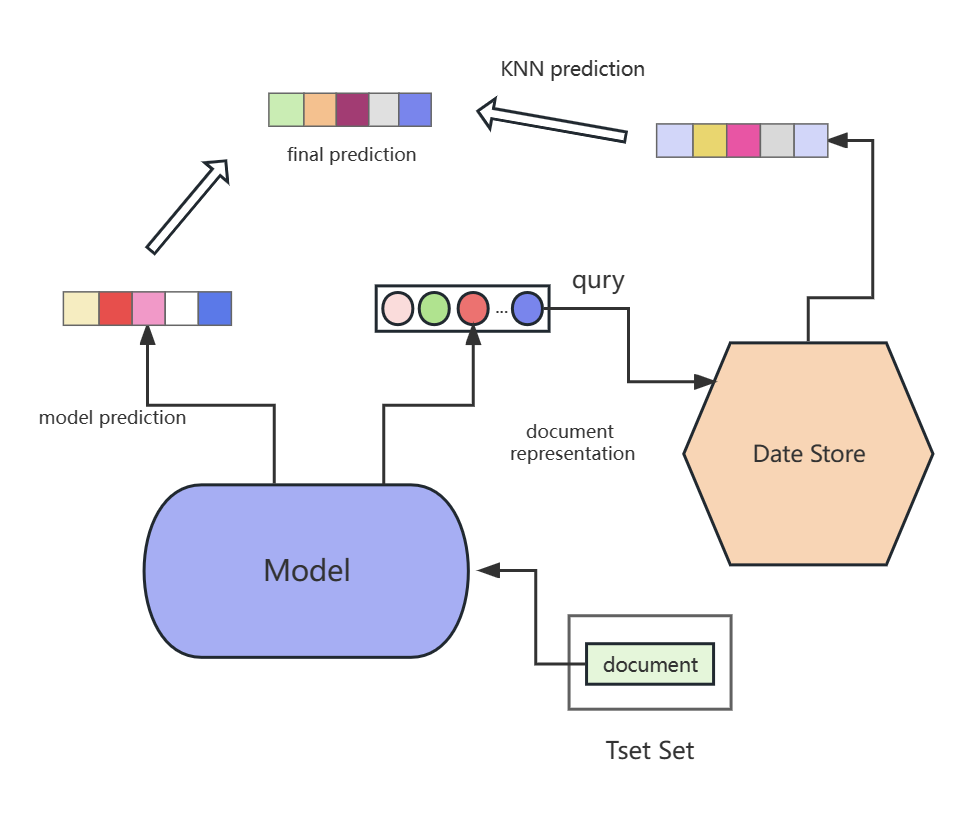}
    \caption{The Overall prediction process with KNN mechanism}
    \label{3}
\end{figure}

\section{Experimental Setup}
In the dataset, many pronouns in the premise were used such as "this research", "it", "this way" etc, and these pronouns refer to objects contained in the conclusion. Whereas our model tries to establish the attention scores of different semantic components of a document for a specific label, it is clear that the presence of these words with unclear denotations affects the attention results. In addition, stance toward conclusions also influence value judgments. Therefore, we use a simple strategy: combining the three parts of conclusion, preference, and stance into a sentence that conforms to natural language conventions. Specifically, the data were preprocessed uniformly and simply, and the input structure was: "I agree (disagree) that" + conclusion content + ", because" + premise content.

We use the Roberta model\cite{liu_roberta_2019} as the base pre-trained model to obtain word representation. Experimenting with our architecture on the base model. And we use the K-fold cross-validation method. We merge the training and validation sets and then randomly divide them into six copies. We perform the training process six times, each time using 5/6 of the data as the training set and 1/6 of the data as the validation. During the training process, the best-trained model for each fold is saved, and the average output probability of all models is taken as the final prediction score. 

\section{Results}

On the leaderboard of the TIRA, our method achieved an macro-F1 score of 0.53 and ranks fourth, while the baseline which use a bert model\cite{devlin_bert_2019} achieved 0.42, with the best result for the whole competition being 0.56. In addition, our model also achieved an F1-score of 0.32 on each of the two test sets, Nahj al-Balagha and New York Times. This effect is relatively high among all participating teams, which fully demonstrates the robustness and stability of our approach.\footnote{Find it from Appendix}

To illustrate the effectiveness of our architecture, we conducted ablation experiments. The ablation experiments evaluate the performance effect of the model directly on the validation set merged with the dataset from Zhihu. In the ablation experiments, we did not use the strategy of k-fold cross-validation. The results of the ablation experiment are shown in Table \ref{table2-results},
\begin{table}
\centering\small%
\setlength{\tabcolsep}{2.5pt}%
\begin{tabular}{@{}llllll@{}}
\toprule
\textbf{model/score} & \textbf{precise} & \textbf{recall} & \textbf{F1} \\ \midrule
baseline             & 0.474            & 0.572           & 0.518       \\
multi-attention       & 0.475            & 0.579           & 0.522       \\
LSAN                 & 0.472            & 0.580           & 0.520       \\
baseline+KNN         & 0.462            & 0.603           & 0.523       \\
multi-attention+KNN   & 0.482            & 0.577           & 0.525       \\ \bottomrule
\end{tabular}
\caption{Ablation Experiment Results}
\label{table2-results}
\end{table}
which shows all strategy results with Precision, Recall, and marco-F1. We show the results for each method using the average results from three runs. At first, we use the word vector corresponding to the output of the Roberta model [CLS] as the representation vector of the document to connect the classifier as the baseline. We then compared the effect of baseline with LSAN\cite{xiao_label-specific_2019}, just use multi-attention mechanism, and the effect of removing the multi-headed attention mechanism part. As can be seen in the table, after using the multi-headed attention mechanism, the marco-F1 value improves by about 0.3\% compared to the baseline model, while the LSAN mechanism get 0.2\% improvement on F1-score. And after adding the KNN mechanism augmented with contrastive learning alone, the marco-F1-score is improved by about 0.4\%. In the case of the full model, the marco-F1-score improves by about 0.7\% compared to the baseline. This result is within our expectation and illustrates the effectiveness of our method. Then in order to increase the stability and robustness, and to avoid overfitting generation, we use the K-fold cross-validation method, so that our experimental results can be shown relatively stable, which leads to an improvement of about 0.4 percentage points in the F1-scores.
\section{Conclusion}
We propose a multi-label text classification model using a label-specific multi-headed attention mechanism. Compared to previous models of attention mechanisms, the use of multi-headed attention enables specific labels to focus on different semantic components of the document more effectively. Besides, we use the KNN mechanism to exploit the instance information in the training set. We then perform ablation experiments on our architecture to analyze the role of each part and demonstrate the superiority of using a multi-headed attention mechanism.
\clearpage
\bibliography{custom}
\bibliographystyle{acl_natbib}

\clearpage
\appendix
\section{Appendix}
\begin{table}[h]
\centering\small%
\setlength{\tabcolsep}{2.5pt}%
\scalebox{0.7}{
  \begin{tabular}{@{}ll@{\hspace{10pt}}c@{\hspace{5pt}}cccccccccccccccccccccc@{}}
\toprule
\multicolumn{2}{@{}l}{\bf Test set / Approach} & \bf All & \rotatebox{90}{\bf Self-direction: thought} & \rotatebox{90}{\bf Self-direction: action} & \rotatebox{90}{\bf Stimulation} & \rotatebox{90}{\bf Hedonism} & \rotatebox{90}{\bf Achievement} & \rotatebox{90}{\bf Power: dominance} & \rotatebox{90}{\bf Power: resources} & \rotatebox{90}{\bf Face} & \rotatebox{90}{\bf Security: personal} & \rotatebox{90}{\bf Security: societal} & \rotatebox{90}{\bf Tradition} & \rotatebox{90}{\bf Conformity: rules} & \rotatebox{90}{\bf Conformity: interpersonal} & \rotatebox{90}{\bf Humility} & \rotatebox{90}{\bf Benevolence: caring} & \rotatebox{90}{\bf Benevolence: dependability} & \rotatebox{90}{\bf Universalism: concern} & \rotatebox{90}{\bf Universalism: nature} & \rotatebox{90}{\bf Universalism: tolerance} & \rotatebox{90}{\bf Universalism: objectivity} \\
\midrule
\multicolumn{2}{@{}l}{\emph{Main}} \\
& \textcolor{gray}{Best per category} & \textcolor{gray}{.59} & \textcolor{gray}{.61} & \textcolor{gray}{.71} & \textcolor{gray}{.39} & \textcolor{gray}{.39} & \textcolor{gray}{.66} & \textcolor{gray}{.50} & \textcolor{gray}{.57} & \textcolor{gray}{.39} & \textcolor{gray}{.80} & \textcolor{gray}{.68} & \textcolor{gray}{.65} & \textcolor{gray}{.61} & \textcolor{gray}{.69} & \textcolor{gray}{.39} & \textcolor{gray}{.60} & \textcolor{gray}{.43} & \textcolor{gray}{.78} & \textcolor{gray}{.87} & \textcolor{gray}{.46} & \textcolor{gray}{.58} \\
& \textcolor{gray}{Best approach} & \textcolor{gray}{.56} & \textcolor{gray}{.57} & \textcolor{gray}{.71} & \textcolor{gray}{.32} & \textcolor{gray}{.25} & \textcolor{gray}{.66} & \textcolor{gray}{.47} & \textcolor{gray}{.53} & \textcolor{gray}{.38} & \textcolor{gray}{.76} & \textcolor{gray}{.64} & \textcolor{gray}{.63} & \textcolor{gray}{.60} & \textcolor{gray}{.65} & \textcolor{gray}{.32} & \textcolor{gray}{.57} & \textcolor{gray}{.43} & \textcolor{gray}{.73} & \textcolor{gray}{.82} & \textcolor{gray}{.46} & \textcolor{gray}{.52} \\
& \textcolor{gray}{BERT} & \textcolor{gray}{.42} & \textcolor{gray}{.44} & \textcolor{gray}{.55} & \textcolor{gray}{.05} & \textcolor{gray}{.20} & \textcolor{gray}{.56} & \textcolor{gray}{.29} & \textcolor{gray}{.44} & \textcolor{gray}{.13} & \textcolor{gray}{.74} & \textcolor{gray}{.59} & \textcolor{gray}{.43} & \textcolor{gray}{.47} & \textcolor{gray}{.23} & \textcolor{gray}{.07} & \textcolor{gray}{.46} & \textcolor{gray}{.14} & \textcolor{gray}{.67} & \textcolor{gray}{.71} & \textcolor{gray}{.32} & \textcolor{gray}{.33} \\
& \textcolor{gray}{1-Baseline} & \textcolor{gray}{.26} & \textcolor{gray}{.17} & \textcolor{gray}{.40} & \textcolor{gray}{.09} & \textcolor{gray}{.03} & \textcolor{gray}{.41} & \textcolor{gray}{.13} & \textcolor{gray}{.12} & \textcolor{gray}{.12} & \textcolor{gray}{.51} & \textcolor{gray}{.40} & \textcolor{gray}{.19} & \textcolor{gray}{.31} & \textcolor{gray}{.07} & \textcolor{gray}{.09} & \textcolor{gray}{.35} & \textcolor{gray}{.19} & \textcolor{gray}{.54} & \textcolor{gray}{.17} & \textcolor{gray}{.22} & \textcolor{gray}{.46} \\
& our model & .53 & .53 & .70 & .26 & .29 & .60 & .45 & .54 & .31 & .77 & .65 & .58 & .60 & .51 & .16 & .59 & .42 & .73 & .85 & .43 & .55 \\
\addlinespace
\multicolumn{2}{@{}l}{\emph{Nahj al-Balagha}} \\
& \textcolor{gray}{Best per category} & \textcolor{gray}{.48} & \textcolor{gray}{.18} & \textcolor{gray}{.49} & \textcolor{gray}{.50} & \textcolor{gray}{.67} & \textcolor{gray}{.66} & \textcolor{gray}{.29} & \textcolor{gray}{.33} & \textcolor{gray}{.62} & \textcolor{gray}{.51} & \textcolor{gray}{.37} & \textcolor{gray}{.55} & \textcolor{gray}{.36} & \textcolor{gray}{.27} & \textcolor{gray}{.33} & \textcolor{gray}{.41} & \textcolor{gray}{.38} & \textcolor{gray}{.33} & \textcolor{gray}{.67} & \textcolor{gray}{.20} & \textcolor{gray}{.44} \\
& \textcolor{gray}{Best approach} & \textcolor{gray}{.40} & \textcolor{gray}{.13} & \textcolor{gray}{.49} & \textcolor{gray}{.40} & \textcolor{gray}{.50} & \textcolor{gray}{.65} & \textcolor{gray}{.25} & \textcolor{gray}{.00} & \textcolor{gray}{.58} & \textcolor{gray}{.50} & \textcolor{gray}{.30} & \textcolor{gray}{.51} & \textcolor{gray}{.28} & \textcolor{gray}{.24} & \textcolor{gray}{.29} & \textcolor{gray}{.33} & \textcolor{gray}{.38} & \textcolor{gray}{.26} & \textcolor{gray}{.67} & \textcolor{gray}{.00} & \textcolor{gray}{.36} \\
& \textcolor{gray}{BERT} & \textcolor{gray}{.28} & \textcolor{gray}{.14} & \textcolor{gray}{.09} & \textcolor{gray}{.00} & \textcolor{gray}{.67} & \textcolor{gray}{.41} & \textcolor{gray}{.00} & \textcolor{gray}{.00} & \textcolor{gray}{.28} & \textcolor{gray}{.28} & \textcolor{gray}{.23} & \textcolor{gray}{.38} & \textcolor{gray}{.18} & \textcolor{gray}{.15} & \textcolor{gray}{.17} & \textcolor{gray}{.35} & \textcolor{gray}{.22} & \textcolor{gray}{.21} & \textcolor{gray}{.00} & \textcolor{gray}{.20} & \textcolor{gray}{.35} \\
& \textcolor{gray}{1-Baseline} & \textcolor{gray}{.13} & \textcolor{gray}{.04} & \textcolor{gray}{.09} & \textcolor{gray}{.01} & \textcolor{gray}{.03} & \textcolor{gray}{.41} & \textcolor{gray}{.04} & \textcolor{gray}{.03} & \textcolor{gray}{.23} & \textcolor{gray}{.38} & \textcolor{gray}{.06} & \textcolor{gray}{.18} & \textcolor{gray}{.13} & \textcolor{gray}{.06} & \textcolor{gray}{.13} & \textcolor{gray}{.17} & \textcolor{gray}{.12} & \textcolor{gray}{.12} & \textcolor{gray}{.01} & \textcolor{gray}{.04} & \textcolor{gray}{.14} \\
& our model & .32 & .06 & .39 & .31 & .44 & .66 & .10 & .33 & .59 & .41 & .16 & .45 & .24 & .16 & .31 & .35 & .20 & .25 & .25 & .00 & .28 \\
\addlinespace
\multicolumn{2}{@{}l}{\emph{New York Times}} \\
& \textcolor{gray}{Best per category} & \textcolor{gray}{.50} & \textcolor{gray}{.50} & \textcolor{gray}{.22} & \textcolor{gray}{.00} & \textcolor{gray}{.03} & \textcolor{gray}{.54} & \textcolor{gray}{.40} & \textcolor{gray}{.00} & \textcolor{gray}{.50} & \textcolor{gray}{.59} & \textcolor{gray}{.52} & \textcolor{gray}{.22} & \textcolor{gray}{.33} & \textcolor{gray}{1.00} & \textcolor{gray}{.57} & \textcolor{gray}{.33} & \textcolor{gray}{.40} & \textcolor{gray}{.62} & \textcolor{gray}{1.00} & \textcolor{gray}{.03} & \textcolor{gray}{.46} \\
& \textcolor{gray}{Best approach} & \textcolor{gray}{.34} & \textcolor{gray}{.22} & \textcolor{gray}{.22} & \textcolor{gray}{.00} & \textcolor{gray}{.00} & \textcolor{gray}{.48} & \textcolor{gray}{.40} & \textcolor{gray}{.00} & \textcolor{gray}{.00} & \textcolor{gray}{.53} & \textcolor{gray}{.44} & \textcolor{gray}{.00} & \textcolor{gray}{.18} & \textcolor{gray}{1.00} & \textcolor{gray}{.20} & \textcolor{gray}{.12} & \textcolor{gray}{.29} & \textcolor{gray}{.55} & \textcolor{gray}{.33} & \textcolor{gray}{.00} & \textcolor{gray}{.36} \\
& \textcolor{gray}{BERT} & \textcolor{gray}{.24} & \textcolor{gray}{.00} & \textcolor{gray}{.00} & \textcolor{gray}{.00} & \textcolor{gray}{.00} & \textcolor{gray}{.29} & \textcolor{gray}{.00} & \textcolor{gray}{.00} & \textcolor{gray}{.00} & \textcolor{gray}{.53} & \textcolor{gray}{.43} & \textcolor{gray}{.00} & \textcolor{gray}{.00} & \textcolor{gray}{.00} & \textcolor{gray}{.57} & \textcolor{gray}{.26} & \textcolor{gray}{.27} & \textcolor{gray}{.36} & \textcolor{gray}{.50} & \textcolor{gray}{.00} & \textcolor{gray}{.32} \\
& \textcolor{gray}{1-Baseline} & \textcolor{gray}{.15} & \textcolor{gray}{.05} & \textcolor{gray}{.03} & \textcolor{gray}{.00} & \textcolor{gray}{.03} & \textcolor{gray}{.28} & \textcolor{gray}{.03} & \textcolor{gray}{.00} & \textcolor{gray}{.05} & \textcolor{gray}{.51} & \textcolor{gray}{.20} & \textcolor{gray}{.00} & \textcolor{gray}{.07} & \textcolor{gray}{.03} & \textcolor{gray}{.12} & \textcolor{gray}{.12} & \textcolor{gray}{.26} & \textcolor{gray}{.24} & \textcolor{gray}{.03} & \textcolor{gray}{.03} & \textcolor{gray}{.33} \\
& our model & .32 & .22 & .12 & .00 & .00 & .47 & .29 & .00 & .22 & .53 & .41 & .00 & .32 & .50 & .15 & .21 & .40 & .56 & .33 & .00 & .38 \\
\bottomrule
\end{tabular}
}
\caption{Achieved macro-F1-score of our team per test dataset, for each of the 20 value categories.}
\label{table1-results}
\end{table}
\end{document}